%% file: Hoffmann_habilitation_2021.tex
\def\output_for_a5{0}
\if\output_for_a51
\documentclass{ifi_thesis_for_a5}
\else
\documentclass{ifi_thesis}
\fi

\usepackage{amsthm}
\usepackage{color}
\usepackage[colorlinks=true]{hyperref}
\hypersetup{linkcolor=black,citecolor=black,urlcolor=black}
\usepackage{pdfpages}
\usepackage{epstopdf}
\usepackage[T1]{fontenc} 
\usepackage{rotating}

\thesisType{Habilitation Thesis}
\date{\today}
\title{Learning body models: from humans to humanoids}
\author{Matej Hoffmann}
\home{Prague}
\country{Czech Republic}
\email{matej.hoffmann@fel.cvut.cz}
\begindate{<start date of thesis>}
\enddate{<end date of thesis>}
\usepackage{draftwatermark}

\begin{document}

\SetWatermarkAngle{0}
\SetWatermarkColor{black}
\SetWatermarkLightness{0.5}
\SetWatermarkFontSize{10pt}
\SetWatermarkVerCenter{30pt}
\SetWatermarkText{\parbox{30cm}{%
\centering This habilitation thesis was accepted at FEE, CTU in Prague in October 2022. This document does not contain the appendices.\\
\centering Full version with appendices available at \url{https://dspace.cvut.cz/handle/10467/98714}.\\
\centering Cite as: Hoffmann, M. (2021), 'Learning body models: from humans to humanoids', \\
\centering Habilitation thesis, Faculty of Electrical Engineering, Czech Technical University in Prague.\\
}}

\input{./ifi_pages.tex}


\begin{acknowledgements}
First, I would like to express thanks to my Humanoid and cognitive robotics research group at the Department of Cybernetics, Faculty of Electrical Engineering, Czech Technical University in Prague. The contributions of Karla Štěpánová, Zdeněk Straka, Petr Švarný, Filipe Gama, Jakub Rozlivek, and Lukáš Rustler constitute an important part of the thesis.  Second, since I joined the Department of Cybernetics in 2017, the Department has always provided me with exceptional support. I would like to thank the Dept. Heads (originally Jan Kybic, later Tomáš Svoboda), Kristina Lukešová for support in HR and other matters, and Hana Pokorná in procurement. Special thanks go to Tomáš Svoboda for an absolutely unique combination of support, independence, and mentorship that he has granted me and my growing research group, and Petra Ivaničová who has been taking care of the budget of my research group and all administrative issues related to grant applications and management. Third, I want to thank to the research groups where I studied and performed research in the past: the Artificial Intelligence Laboratory, Department of Informatics\footnote{The template used for this thesis is one of Department of Informatics.}, University of Zurich (2006-2013) and Italian Institute of Technology (IIT, 2013-2016). The PhD study at the Zurich AI Lab under Rolf Pfeifer's supervision has provided a foundation for my career. Postdoctoral research at IIT and collaboration in particular with Alessandro Roncone and Ugo Pattacini constitutes another foundation for my current research. Fourth, I want to thank my external collaborators, in particular Kevin O'Regan, Jeffrey J. Lockman, Lisa Chinn, and Daniela Corbetta. Finally, I need to thank my wife Veronika, my daughters Anna and Tereza, and my parents. 
\\ \\
The research reviewed in this thesis would not have been possible without the generous funding from:
\begin{itemize}
 \item 2020-2024 -- Czech Science Foundation (GA CR): ``Whole-body awareness for safe and natural interaction: from brains to collaborative robots''. Project no. 20-24186X.  
 \item 2017-2019 -- Czech Science Foundation (GA CR): ``Robot self-calibration and safe physical human-robot interaction inspired by body representations in
primate brains''. Project no. 17-15697Y. 
 \item 2018-2022 -- The Ministry of Education, Youth and Sports of the Czech Republic, OP VVV funded project CZ.02.1.01/0.0/0.0/16\_019/0000765 ``Research Center for Informatics''.
 \item 2014-2016 -- Marie Curie Intra-European Fellowship (FP7-PEOPLE-2013-IEF 625727): ``iCub Body Schema''.
\end{itemize}
\end{acknowledgements}

\begin{abstract}
Humans and animals excel in combining information from multiple sensory modalities, controlling their complex bodies, adapting to growth, failures, or using tools. These capabilities are also highly desirable in robots. They are displayed by machines to some extent. Yet, the artificial creatures are lagging behind. The key foundation is an internal representation of the body that the agent---human, animal, or robot---has developed. The mechanisms of operation of body models in the brain are largely unknown and even less is known about how they are constructed from experience after birth. In collaboration with developmental psychologists, we conducted targeted experiments to understand how infants acquire first ``sensorimotor body knowledge''. These experiments inform our work in which we construct embodied computational models on humanoid robots that address the mechanisms behind learning, adaptation, and operation of multimodal body representations. At the same time, we assess which of the features of the ``body in the brain'' should be transferred to robots to give rise to more adaptive and resilient, self-calibrating machines. We extend traditional robot kinematic calibration focusing on self-contained approaches where no external metrology is needed: self-contact and self-observation. Problem formulation allowing to combine several ways of closing the kinematic chain simultaneously is presented, along with a calibration toolbox and experimental validation on several robot platforms. Finally, next to models of the body itself, we study peripersonal space---the space immediately surrounding the body. Again, embodied computational models are developed and subsequently, the possibility of turning these biologically inspired representations into safe human-robot collaboration is studied.  
\end{abstract}

\tableofcontents

\mainmatter


\chapter{Introduction}

The main theme in my research has been embodied cognition and its development, how it can be studied using robots, and how inspiration from biology can drive progress in robotic technology. The specific focus is on representations of the body and the space around it---peripersonal space. This habilitation thesis is constituted by a collection of thirteen articles (published or accepted for publication), preceded by a common introductory part that links these works into a coherent whole. 

\section{Synthetic methodology}
\label{sec:synthetic_methodology}
\begin{figure}[htbp]
  \centerline{\includegraphics[width=0.8\textwidth]{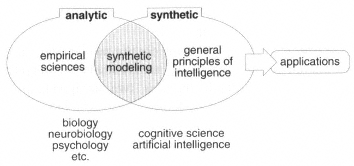}}
  \caption{\textbf{Synthetic methodology. Overview of approaches to the study of intelligence.} On the left, we have the empirical sciences like biology, neurobiology, and psychology that mostly follow an analytic approach. In the center, we have the synthetic ones, namely cognitive science and AI, which can either model natural agents (this is called synthetic modeling, the shaded area) or alternatively can simply explore issues in the study of intelligence without necessarily being concerned about natural systems. From this latter activity, industrial applications can be developed. Figure and caption from \cite[Chapter 1]{Pfeifer2001}.}
  \label{fig:synthetic_methodology}
\end{figure}

The methodology employed in my work is the so-called \textit{synthetic methodology}, or ``understanding by building'' \cite[Chapter 1]{Pfeifer2001} \cite[Chapter 3]{PfeiferBongard2007}. This approach can be applied to study phenomena from many disciplines like biology, psychology, or physics. As a matter of fact, many sciences that were traditionally mostly analytical are becoming more synthetic, employing computer simulations, for example. Here we focus on the understanding of intelligence or cognition in particular. The synthetic methodology, schematically illustrated in Fig.~\ref{fig:synthetic_methodology}, unites the following three goals \cite[Chapter 3]{PfeiferBongard2007}: 
\begin{enumerate}
 \item understanding natural forms of intelligence
 \item abstracting general principles of intelligent behavior 
 \item building intelligent artifacts 
\end{enumerate}

A very influential paradigm in the study of cognition and intelligence was cognitivism (e.g., \cite{Fodor1975,Pylyshyn1984}), whereby thinking was understood as a result of computation over symbols that represent the world. Such computation in the form of automatic manipulation of abstract symbols was at the heart of the so-called ``Good Old-Fashioned Artificial Intelligence'' (GOFAI) \cite{Haugeland1985}. In fact, the computer was not only a tool for modeling but became a metaphor for the mind. 
Cognitivism and GOFAI were thus employing the full synthetic methodology in (1) applying the computer metaphor to model the mind; (2) abstracting general principles and developing algorithms for planning, reasoning, etc.; (3) developing applications like chess-playing programs, expert systems etc.  

\subsection{Embodied computational modeling of cognition}

While sucessful in the abstract domains, GOFAI faced severe difficulties when controlling robots that had to interact with the physical world in real time. Rodney Brooks has openly attacked the GOFAI position in the seminal articles ``Intelligence without representation'' \cite{brooks1991intelligenceWithoutRepresentation} and ``Intelligence without reason'' \cite{brooks1991intelligenceWithoutReason}. Through building robots that interact with the real world, such as insect robots \cite{Brooks1989}, he realized that ``when we examine very simple level intelligence we find that explicit representations and models of the world simply get in the way. It turns out to be better to use the world as its own model.'' \cite{brooks1991intelligenceWithoutRepresentation} The thesis that intelligent behavior emerges from the dynamic interplay of brain, body and environment has also been articulated by the notion of \textit{embodiment} (e.g., \cite{Pfeifer2001,PfeiferBongard2007}) that also started to challenge cognitivism as the dominant paradigm in cognitive science. Different variants of embodied cognition theories have been articulated like the dynamic systems approach to the development of cognition and action \cite{ThelenSmith1994}, grounded cognition \cite{Barsalou2008}, sensorimotor contingency theory \cite{O'Regan2001}, or enaction \cite{Varela1991}. 

Embodied cognition theories stressed the constitutive role of the body and action for even high-level abstract thinking. Given the importance of the body and closed sensorimotor loops, the analytical and divide-and-conquer approach of empirical sciences studying specific phemonena in isolation is more difficult to apply. Instead, a more holistic approach should be adopted. Applying the synthetic methodology to embodied cognition means not only modeling cognitive processes but requires building complete artifacts interacting with the environment. Grey Walter \cite{Walter1953} was the pioneer of this approach already before the era of computers and cognitivism building electronic machines with a minimal ``brain'' that displayed phototactic-like behavior. This was picked up by Valentino Braitenberg who built a whole series of two-wheeled vehicles of increasing complexity, as summarized in ``Vehicles -- Experiments in Synthetic Psychology'' \cite{Braitenberg1986}. Already the most primitive ones, in which sensors are directly connected to motors (exciting or inhibiting them), displayed sophisticated behaviors. The obvious tools for this type of modeling became robots, giving rise to cognitive developmental robotics \cite{Asada2009,Cangelosi2015}. Whenever the emphasis is on the compatibility with available knowledge on the anatomy and physiology of the brain structures underlying the behaviors of interest, the label neurorobotics is also used (e.g., \cite{vanDerSmagt2016neurorobotics}).

Let us first focus on the goal of understanding cognition and its development through syntethic modeling. The perspective of some of the leading researchers in this field is offered in ``The mechanics of embodiment: a dialog on embodiment and computational modeling'' \cite{Pezzulo2011}. 
This approach is also articulated by Caligiore et al.~\cite{caligiore2010tropicals,caligiore2014integrating} and called \textit{computational embodied neuroscience}, with the following goals or constraints for the model:
\begin{itemize}
   \item accurately reproduce the behaviors observed in specific psychological experiments
   \item reproduce the learning processes alongside the final behavior
   \item use architectures and algorithms constrained by neuroscientific evidence
   \item the model should control an embodied agent
\end{itemize}
Certain aspects of a natural system are studied, abstracted, and finally reproduced in an artificial system, which is then subject to investigations. The behavior as well as developmental and learning processes leading to the final behavior of the artificial system are compared with the original natural system, serving as a model of the phenomenon. Not only the overt behavior but also the inner workings responsible for it are compared. The key question is what can embodied cognition learn from synthetic modeling. The key benefits, inspired by Cangelosi and Parisi \cite{cangelosi2002computer}, are:
\begin{enumerate}
 \item Explicit and detailed expression of theory. If the theory is ``embedded'' in a robot, it has to be explicit, detailed, and complete because otherwise it could not be tested in the robot. 
 \item Alternative validation. Any theory should ultimately be validated by comparing it to the biological or psychological phenomenon of interest. However, after implementing it in a robot, one can verify if it generates the expected behavior---for example, whether a machine built according to a model of walking actually walks. Often, this will not be the case and several iterations improving the theory will follow.
 \item New and detailed predictions. The behavior of the robot may not match with the original expectations or may display new unexpected characteristics that can be compared with existing empirical data or lead to the design of new experiments.
 \item Virtual experimental laboratories. With embodied computational models, one typically has access to all variables, including internal variables (raw and preprocessed sensory values, internal states, motor commands, etc.). In addition, it is often relatively easy to manipulate these variables and thus simulate experiments that would otherwise be impossible or ethically unacceptable to perform (e.g., simulating sensory or motor impairments).
\end{enumerate}

My account of ``Robots as powerful allies for the study of embodied cognition from the bottom up'' is provided in \cite{HoffmannPfeifer2018} and included in Appendix~\ref{appendixA:robots_for_embodied_cognition}. 

\subsection{Embodied computational modeling of body representations}

Synthetic modeling can be also productively applied to the study of body and peripersonal space representations. Since the focus is on models of the body, the need for embodied models is even stronger. Humanoid robots possess morphologies---physical characteristics as well as sensory and motor apparatus---that are in some respects akin to human bodies and can thus be used to expand the domain of computational modeling by anchoring it to the physical environment and a physical body and allowing for instantiation of complete sensorimotor loops. The iCub humanoid robot \cite{Metta2010} stands out in this respect: it has anthropomorphic proportions modeled after a 4-year old child and a corresponding set of sensory modalities including binocular vision, hearing, vestibular (inertial) sensing, proprioception, and the recent important addition: touch \cite{Bartolozzi2016}, making it the ideal platform to model the multimodal contributions to body representations. Various surveys on body schema in robotics \cite{Hoffmann2010}, exploration, body representations and internal simulation \cite{Schillaci2016}, ``synthetic psychology of the self'' \cite{prescott2019synthetic} and robot models to study the (en)active self \cite{Lanillos2017enactive,nguyen2021sensorimotor} have been published. 

Remarkable demonstrations of this approach are the models of foetus development from the laboratory of Yasuo Kuniyoshi at the University of Tokyo (e.g., \cite{Yamada2016}). They also illustate the potential of these models as virtual experimental laboratories, showing for example the effect of non-uniform versus uniform tactile sensor distribution \cite{Mori2010} or uterine versus extrauterine environment \cite{Yamada2016}.

Finally, it is important to note that not only is embodiment constitutive for the nature and operation of body models in the brain (and hence humanoid robots should be used to study body representations), but also, paraphrasing Brooks \cite{brooks1991intelligenceWithoutRepresentation} the body may sometimes be its own best model. In other words, the physical body may be used directly, without having to have it represented in the brain (see ``Embodying the mind and representing the body'' \cite{Alsmith2012} or ``body-in-the-brain versus body-in-the-world'' \cite[p. XV-XVI]{ataria2021body} for a deeper discussion). Thus, with computational modeling only, one can hardly adopt other than the representationalist approach to body maps \cite[Chapter 5]{deVignemont2018mind}. Instead, with robots, the sensorimotor approach to body know-how (e.g., \cite{jacquey2020development}) can also be modeled.  

\subsection{Abstracting general principles of learning body models and applications}
Apart from synthetic modeling of the mechanims of the ``body in the brain'', robots autonomously learning their body maps serve also abstracting general principles and even applications (see Fig.~\ref{fig:synthetic_methodology} again for the different aspects of the synthetic methodology). General characteristics of body models in animals, humans, and robots are derived in \cite{Hoffmann_BodyModelsOUP_2021}, included in Appendix~\ref{appendixB:body_models}. The applications of robots learning their body models are in automatic self-contained robot self-calibration and they are reviewed in \cite{Hoffmann_EncRobotics_2021} and included in Appendix~\ref{appendixC:self-calibration}.

\section{Body models in humans, animals, and robots}
A theoretical account contrasting the body representations in the octopus, humans, and robots is presented in \cite{Hoffmann_BodyModelsOUP_2021}, included in Appendix~\ref{appendixB:body_models}. On one hand, the example of the models of the iCub robot (like forward and inverse kinematics, inverse dynamics, etc.) and how it reaches for a tactile stimulus in \cite{Roncone_ICRA_2014} serves as an intuition pump to analyze body models in animals and humans. On the other hand, what is known about how animals use and represent their bodies, including the imporant aspect that behavior generation is in the brain-body-environment coupling and hence not all features need to be modeled, is used to inspire the design of future generations of machines that would inherit some of the capacities demonstrated by animals like resilience.

\section{Thesis overview}
This thesis is a collection of published works, complemented by relatively brief texts that explain the relationships between the individual articles. The story of the thesis is largely told in three book chapters and the reader is advised to consult them. In \cite{HoffmannPfeifer2018}, Appendix~\ref{appendixA:robots_for_embodied_cognition}, the methodology of using humanoid robots to study embodied cognition is explained and illustrated on examples. In \cite{Hoffmann_BodyModelsOUP_2021}, Appendix~\ref{appendixB:body_models}, body models in humans, animals, and robots are contrasted. In \cite{Hoffmann_EncRobotics_2021}, Appendix~\ref{appendixC:self-calibration}, an overview of biologically inspired robot body models and self-calibration is provided for a robotics encyclopedia. All three articles were prepared for a wider, often interdisciplinary, audience, and they are thus accessible even for non-experts in a specific field.

\begin{sidewaysfigure}[htbp]
  \centerline{\includegraphics[width=0.95\textwidth]{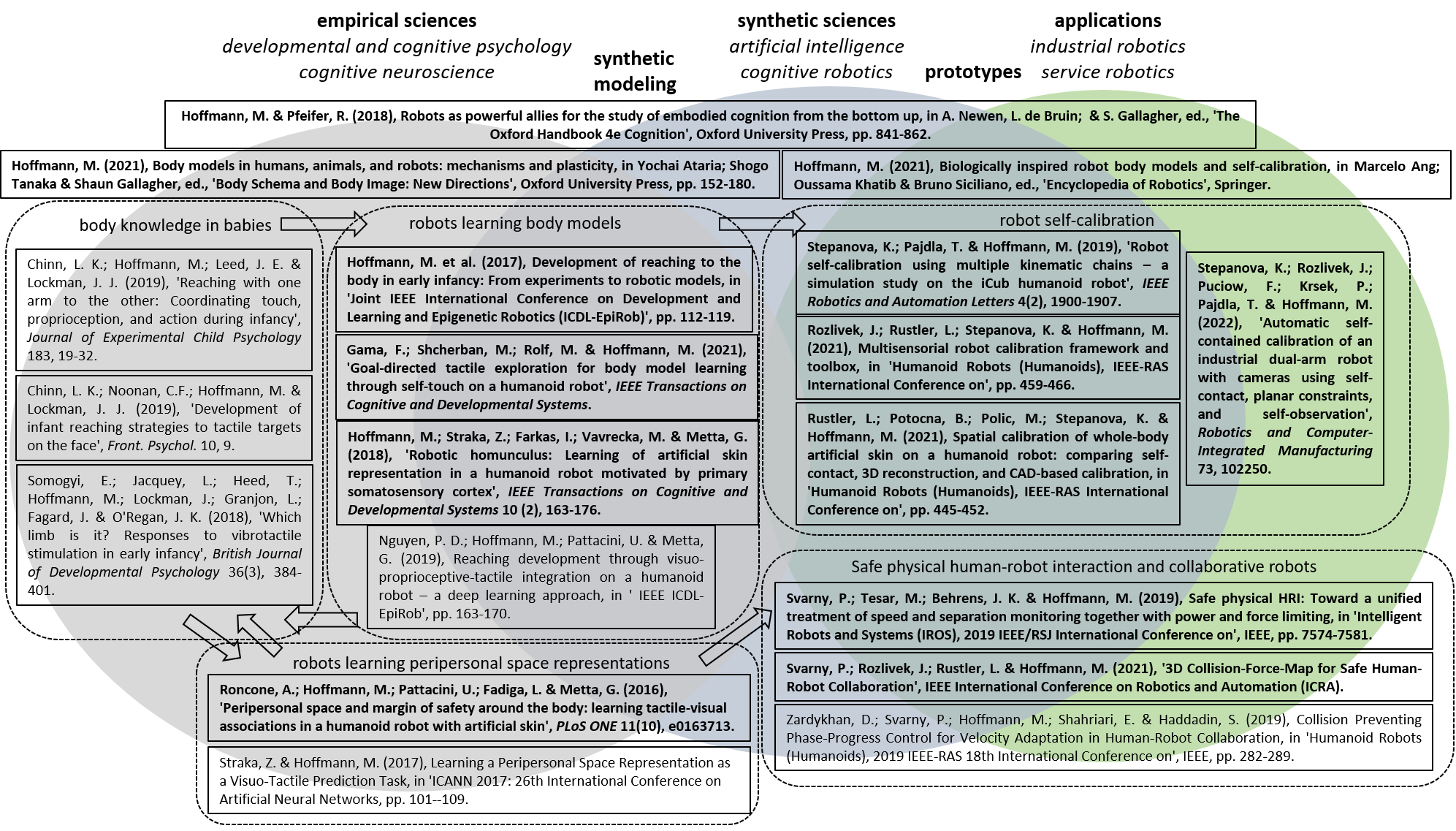}}
  \caption{\textbf{Habilitation Overview.} Schematic visualization of the articles composing the thesis, grouped according to common themes, and overlayed on the synthetic methodology (analogous to Fig.~\ref{fig:synthetic_methodology}). The articles in bold letters are included in the thesis appendices.}
  \label{fig:hab_overview}
\end{sidewaysfigure}

The rest of the thesis brings together articles with a more technical content and groups them as follows. In Chapter~\ref{chapter:body_knowledge_babies}, an overview of related research on the development of body representations in children is provided, along with my own contributions through collaborations with developmental psychologists. Chapter~\ref{chapter:robots_learning_body_models} describes biologically motivated case studies on robots learning body models. Chapter~\ref{chapter:robots_learning_pps} focuses on robots learning about the space surrounding their bodies, so called peripersonal space. Chapter~\ref{chapter:self-calibration} deals with the topic of automatic self-contained robot calibration. Chapter~\ref{chapter:pHRI} brings together works on safety of physical human-robot interaction. The thesis Appendix includes thirteen peer-reviewed published works pertinent to the thesis topic. With one exception \cite{Roncone2016}, I am either the first or the senior author on all these publications---in the latter case, they originate from my research group.

\chapter{Body knowledge in babies}
\label{chapter:body_knowledge_babies}
A wealth of observations has been accumulated about development of the foetus and the infant in the months before and after birth. There is certainly a great deal to learn for the infant in this period, but the assumption in my work is that the, possibly primary, target of what the infant is learning about in this period is its own body. Understanding what  its body parts are, what it can do with them, and what the consequences would be, are essential prerequisites for almost everything that follows. This is the sensorimotor or ecological self \cite{neisser1988five}. The second assumption relates to the way in which the infant learns. In my work, self-touch or haptic exploration of one's own body is considered a key activity that may bootstrap learning about the body as an object in space \cite{Hoffmann_selfTouch_2017}. We assume that motor-proprioceptive-tactile contingencies provide sufficient material for the infant to construct ``functional body knowledge'', or body know-how, allowing her for example to reach for specific body parts when presented with a tactile stimulus. This sensorimotor account is presented in \cite{Hoffmann2017icdl} and included in Appendix~\ref{appendixD:reaching_to_the_body}. The role of vision is increasingly important during postnatal development as the acuity and gaze control improves. Eventually, it is likely that the representations of the body and peripersonal space become predominantly vision-based. However, to what extent, when, and how this occurs still needs to be established.

\section{Spontaneous self-touch}
\footnote{This section is adapted from Section II in \cite{Gama_TCDS_2021}.}
Fetuses initially perform local movements directed to areas of the body most sensitive to touch: the face, but also soles of feet \cite[p.~113-114]{Piontelli2015}. Later, from 26 to 28 weeks of gestational age, they also use the back of the hands and touch other body areas like thighs, legs, and knees \cite[p.~29-30]{Piontelli2015}. In addition, from 19 weeks, fetuses anticipate hand-to-mouth movements \cite{Myowa2006} (the mouth opens prior to contact) and from 22 weeks, the movements seem to show the recognizable form of intentional actions, with kinematic patterns that depend on the goal of the action (toward the mouth vs. toward the eyes) \cite{Zoia2007}. 

Hand-mouth coordination continues to develop after birth \cite{Rochat1993}. Specifically related to body exploration, \cite{Rochat1998} writes: ``By 2-3 months, infants engage in exploration of their own body as it moves and acts in the environment. They babble and touch their own body, attracted and actively involved in investigating the rich intermodal redundancies, temporal contingencies, and spatial congruence of self-perception.'' \cite{DiMercurio2018} followed spontaneous behavior of infants from 3 to 9 weeks; \cite{Thomas2015} from birth to 6 months of age. Their main findings regarding self-touch were:
\begin{itemize}
 \item rostro-caudal progression: head and trunk contacts are more frequent in the beginning, followed by more caudal body locations including hips, then legs, and eventually the feet \cite{Thomas2015} 
 \item contacts are typically made with the ipsilateral hand
 \item \emph{complex touches}, as infants moved their hand while remaining in contact with their body, were frequently observed by \cite{DiMercurio2018} 
\end{itemize}

In summary, infants acquire ample experience of touching their body, which allows for the learning of the first tactile-proprioceptive-motor models of the body. The ability to learn from this experience goes hand in hand with dynamic neural development in this period \cite{Tau2010}; see \cite{Hoffmann_selfTouch_2017} for a review focusing specifically on self-touch. Yet, the behavioral organization of such early tactile exploration is not understood. Are the touches on the body spontaneous or systematic? If there is a particular structure---which seems to be the case \cite{DiMercurio2018,Thomas2015}---what drives this developmental progression?  \cite{Piaget1952} theorized that in newborns, action and perception as well as the ``spaces'' of individual sensory modalities are separated (cf. \cite{vanDerMeer1995} for evidence  that visual and motor modalities are connected early after birth). Until the connections are established, infants explore their environment (and their body) randomly. \cite{Piaget1952} also proposed a pivotal role of repeated movements---\textit{primary circular reactions} directed to learn properties of the body and \textit{secondary circular reactions} driven by the interest on the effects they produce in the environment. However, to discriminate spontaneous contacts from systematic (intrinsically motivated) exploration remains a challenge. 

\section{Localizing touch on the body}
\footnote{The first paragraph is adapted from Section II in \cite{Gama_TCDS_2021}.}
A counterpart to recordings of spontaneous infant behavior is provided by testing how they can reach to targets on their body. Lockman and colleagues performed a series of studies \cite{Chinn_JECP_2019,chinn2021human,Hoffmann2017icdl,Leed2019,Somogyi2018} in which vibrotactile targets (``buzzers'') were attached to infants' body parts and their ability and their way of reaching for the targets were analyzed. Targets above the mouth and on the chin were successfully contacted already from 2 months of age \cite{chinn2021human}, followed by trunk area, legs, hands, other areas on the face (forehead, ears), and elbows (around 9 months) (whole body -- pilot study \cite{Hoffmann2017icdl}; upper body \cite{Leed2019}). For targets on hands and arms, the arm with the buzzer and the contralateral arm reaching for the target often moved simultaneously---the arm with the target actively facilitating the removal \cite{Chinn_JECP_2019}. Regarding looking at the target for locations where this was possible, the infants looked first in 86 of 189 trials (45.5\%), reached first in 27 of 189 trials (14.29\%), and looked and reached simultaneously in 76 of 189 trials (40.21\%). This last strategy significantly increased with age (7 to 21 months).

Another window into the development of tactile localization is provided by experiments that exploit crossed hands or crossed feet postures, often employed to study conflicts between encoding touch in an anatomical versus external frame of reference. In short, when one's left hand is located in the right peripersonal space, there is a conflict between where the hand is normally and currently, which impairs tactile localization (one manifestation is slower response in the temporal order judgement task -- \cite{heed2014using} for a review). \cite{ali2015human} tested two groups of infants---four-month and six-month-olds---by applying tactile stimulation to their feet in a crossed posture. Six-month-olds, like adults, showed a tactile localisation deficit, indicating external spatial coding of touch; in striking contrast, four-month-olds outperformed the older infants showing no crossed-feet deficit, indicating that coding of touch in external space, called also tactile remapping \cite{Heed2015}, develops at around 5 months after birth, possibly in line with the growing importance of vision. In another study, \cite{bremner2008spatial} tested 6.5-month- and 10-month-olds in a crossed and uncrossed hand posture and investigated their manual and looking responses.

\section{My contributions}
The studies briefly reviewed above provide important constraints for the embodied computational modeling work (Chapter~\ref{chapter:robots_learning_body_models}). Some of the authors of the works cited above have become my collaborators (Prof. Jeffrey J. Lockman, Tulane University, USA; Prof. Daniela Corbetta, University of Tennessee, Knoxville, USA; Dr. Kevin O’Regan, University Paris Descartes \& CNRS, Paris, France; Prof. Tobias Heed, Paris Lodron University Salzburg, Austria). I have co-authored the following articles:
\begin{itemize}
 \item Chinn, L. K., Hoffmann, M., Leed, J. E., \& Lockman, J. J. (2019). Reaching with one arm to the other: coordinating touch, proprioception, and action during infancy. \textit{Journal of experimental child psychology}, 183, 19-32. [M.H. author contribution 25\%]
 \item Chinn, L. K., Noonan, C. F., Hoffmann, M., \& Lockman, J. J. (2019). Development of infant reaching strategies to tactile targets on the face. \textit{Frontiers in psychology}, 10, 9. [M.H. author contribution 20\%]
 \item Somogyi, E., Jacquey, L., Heed, T., Hoffmann, M., Lockman, J. J., Granjon, L., Fagard, J., \& O'Regan, J. K. (2018). Which limb is it? Responses to vibrotactile stimulation in early infancy. \textit{British Journal of Developmental Psychology}, 36(3), 384-401. [M.H. author contribution 10\%]
\end{itemize}
These articles are not included in this habilitation which is submitted for the Technical Cybernetics program.

In addition, part II.C of \cite{Hoffmann2017icdl}, Appendix~\ref{appendixD:reaching_to_the_body}, and part II of \cite{Gama_TCDS_2021}, Appendix~\ref{appendixE:goal_directed} contain also sections reporting results of experiments with infants specifically targeting the needs of the modeling endeavour.

\section{Summary}
The observations from the studies on babies provide inputs and constraints for the embodied computational modeling. In line with the synthetic methodology (Section~\ref{sec:synthetic_methodology}), inspiration from biology may contribute to the discovery of general principles that may eventually give rise to useful technology like self-calibrating robots. However, the primary goal is to help uncover the mechanims of the development of body know-how in infants. Collecting unambiguous empirical evidence is complicated by the age of the subjects---infants in the first year are preverbal and one thus cannot easily instruct them and their cooperation during experiments is limited. Similarly, manipulations (e.g., reaching for a tactile stimulus only with a certain limb) are hard to orchestrate and additional setups like motion capture or brain imaging are also almost impossible to arrange. Therefore, robot models that allow arbitrary manipulations and where all internal variables can be accessed provide an indispensable tool.

\chapter{Robots learning body models}
\label{chapter:robots_learning_body_models}

The results of empirical studies from Chapter~\ref{chapter:body_knowledge_babies} flow into embodied computational models on humanoid robots that address the mechanisms of development of reaching and somatosensory perception in early infancy. Humanoid robots with pressure-sensitive electronic skins covering large areas of their bodies provide the right platform for this type of work. There are four articles in this research strand \cite{Hoffmann2017icdl,Gama_TCDS_2021,HoffmannStraka2018,Nguyen_ICDL_2019}, three of which are included in the thesis.

The article \cite{Hoffmann2017icdl}, included in Appendix~\ref{appendixD:reaching_to_the_body}, is a conceptual one, in collaboration with developmental and cognitive psychologists and provides a natural transition from the empirical studies to embodied computational modeling of the development of reaching to the body. 

The second article \cite{Gama_TCDS_2021}, included in Appendix~\ref{appendixE:goal_directed}, focuses on active body exploration and compares different algorithms from the family of intrinsically motivated learning (or artificial curiosity) \cite{Schmidhuber1991,OudeyerKaplan2007,baranes_2013,Baldassarre2013}. This work demonstrates that efficient exploration of the skin space and learning of inverse models is possible. First grounding of the modeling work in experimental data is attempted (study of motor redundancy in infant reaching and link to ``complex touches'' from \cite{DiMercurio2018}). 

The third article \cite{HoffmannStraka2018}, included in Appendix~\ref{appendixF:robot_homunculus}, studies the development of a biologically motivated representation of the robot skin surface: the robot ``tactile homunculus''. The robot is exposed to tactile stimulations on its whole body and the corresponding tactile activations are recorded. These are then fed into a self-organizing (or Kohonen) map algorithm and the representations that emerge are studied. Modifications of the standard algorithm that provide the right constraints to channel learning toward the layout of the neural map observed in primate brains are developed. 

The fourth article \cite{Nguyen_ICDL_2019} studies the development of reaching to objects external to the body. These are perceived visually, but we concentrate on the role of haptic feedback in learning the behavior on the robot. If the object is at first randomly contacted, proprioception provides an alternative to vision to guide subsequent reaching movements. Such ``somatosensory coding of space'' connects reaching to the body with reaching to external objects.

\chapter{Robots learning peripersonal space representations}
\label{chapter:robots_learning_pps}

Next to ``body space'' or ``personal space'', the space immediately surrounding the body is called peripersonal space. There are two related but different meanings associated with peripersonal space (PPS): (i) space immediately surrounding the body and (ii) space that we can act upon / within our reach. Their representations in the brain are realized by fronto-parietal networks, with an important role attributed to bimodal neurons with visuo-tactile receptive fields (RFs) \cite{Clery2015}. The first notion, space surrounding the body, can be pictured as a ``bubble'' around individual body parts and following those body parts in space. It is realized by bimodal neurons with tactile RFs on the skin and visual RFs around. The visual responses appear to be tuned to dynamically approaching objects and their activation is thus anticipatory, predictive of touch in the corresponding skin area. There are also behavioral responses associated with the stimulation of some neurons of this network (squinting, ducking, and withdrawing from the direction of the potential threat). Therefore, this circuit is thought to be responsible for self-defense and maintaining a safety margin around the body \cite{Graziano2006}. Little is known about the development of peripersonal space representations, with some evidence suggesting that even newborns can make sense of multisensory (audio-visual in this case) cue combinations specifying motion with respect to themselves \cite{orioli2018multisensory}.

Taking advantage of humanoid robots with whole-body sensitive skin, we designed a method where the robot learns such a safety margin from experience. The robot records trajectories of approaching objects and if they eventually contact its body as perceived by the artificial skin, it learns the likelihood of such future contact. The objects' positions perceived by the robot by stereo vision are remapped into the reference frames of every tactile sensor (taxel). Every taxel then continuously updates a representation of the probability of objects at a certain distance \cite{Roncone_IROS_2015} or with a certain time to contact \cite{Roncone2016}  
colliding with it. These ``threatened'' taxels can be aggregated and a most likely future collision site on the robot body computed. This location on the robot body can then be connected to a controller that avoids the collision by moving the exposed body part away (video: \url{https://youtu.be/3IaXxNwC_7E}). Alternatively, using the opposite sign for the movement direction, ``whole-body reaching/catching'' can be easily realized. The work \cite{Roncone2016} is included in Appendix~\ref{appendixG:pps}.

We developed an alternative learning algorithm in \cite{Straka2017}, a neural network composed of a Restricted Boltzmann Machine and a feedforward neural network. The former learns in an unsupervised manner to represent position and velocity features of the stimulus. The latter is trained in a supervised way to predict the position of touch (contact). Compared to \cite{Roncone2016}, this model was trained in a simulated environment but it importantly take into account also the the uncertainty of all variables.

\begin{figure}[htbp]
  \centerline{\includegraphics[width=0.8\textwidth]{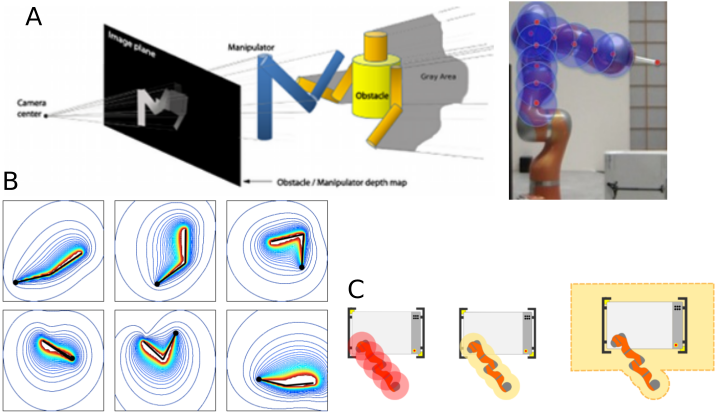}}
  \caption{\textbf{Perirobot space representations.} (A) Depth space approach \cite{Flacco2012}. (B) Kinetostatic danger field \cite{lacevic2010kinetostatic}. (C) Dynamic safety fields \cite{magnanimo2016safeguarding}.}
  \label{fig:perirobot}
\end{figure}

Apart from modeling the operation, development, and adaption of peripersonal space representations in the brain, ``perirobot space'' representations are key for safe human-robot interaction. Appropriate collision avoidance strategies cannot be selected without a representation of the space surrounding the robot. This space can be represented in various forms with regards to the needs of the application or robot platform. Multiple questions come forth in this context, namely the way the space is structured, how it will dynamically adapt and what is its geometry. A key component is the form of the representation of the robot and human body parts, or, in general, the representation of obstacles. Drawing on the results of the computer graphics community (see \cite{jimenez2001_3Dcollision} for a survey), this often takes the form of some collision primitives. These can be simple shapes like spheres \cite{Flacco2012} or more complex meshes \cite{polverini2017computationally} and can differ for the robot and the human. \cite{zanchettin2015safety}  represent robot links as segments and humans as a set of capsules.  These shapes can also have a temporal aspect and represent so-called swept volumes, i.e. zones where the human or robot moved. For safety to be guaranteed, the whole body of both agents should be represented and considering only the robot end-effector or human hands or head does not suffice. Some representations change the volume dynamically based on the robot or human velocity \cite{polverini2017computationally,zanchettin2015safety,lacevic2010kinetostatic,magnanimo2016safeguarding}. The technologies applied to perception can influence the structure of the perirobot space too: Euclidean space can be replaced by a depth space approach to account for the occlusions and specific geometries of the field of view of a RGB-D sensor \cite{Flacco2012,flacco2015depth}. Examples are depicted in Fig.~\ref{fig:perirobot}. The best representation is yet to be found. These considerations tie directly into the topic of safe physical human-robot interaction discussed in Chapter~\ref{chapter:pHRI} and our publication \cite{Svarny_IROS_2019}.

\chapter{Robot self-calibration}
\label{chapter:self-calibration}

As briefly reviewed in Chapter~\ref{chapter:body_knowledge_babies}, newborns have not only limited conceptual and spatial knowledge about their bodies, but also limited ``body know-how''---means to use their bodies for purposeful action or to localize and act on stimuli on their body. They learn the necessary body models in the first years after birth, while continuously incorporating physical body growth as well as maturation of the sensory apparatus (vision in particular). Next to the developmental time scale, human body representations were found to be adaptive (called \textit{plastic} in neuroscience) on much shorter time scales, as demonstrated by the ability to use tools and their incorporation into the body schema \cite{MaravitaIriki2004}, for example. Both capacities---learning and quickly adapting body models---would be highly desired in robots. 

\begin{figure}[htbp]
  \centerline{\includegraphics[width=0.8\textwidth]{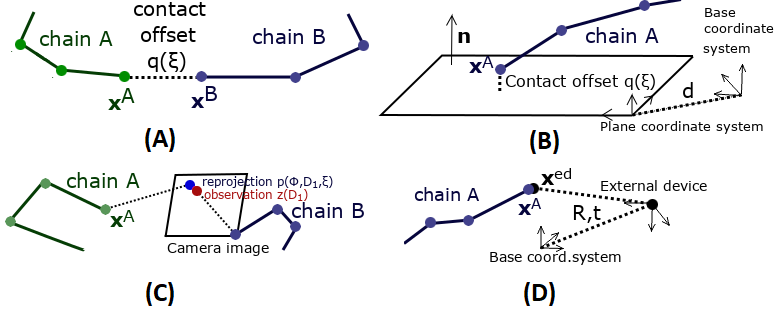}}
  \caption{Schematics of calibration using self-contact (A), contact with a plane (B), self-observation (C), and external device (D). Figure and caption from \cite{Rozlivek_MultiRobotToolbox_2021}.}
  \label{fig:schema_all_contacts}
\end{figure}

Nowadays, humanoid but also other robots come with a rich set of powerful yet inexpensive sensors like cameras, RGB-D cameras, inertial, tactile or force sensors. This opens up the possibility for calibration approaches that are more self-contained, can be performed autonomously and repeatedly by the robot, and that simultaneously estimate the position of the sensors with respect to the robot. The key to self-calibration is \textit{redundancy}. The kinematic chain can be closed exploiting physical contact (aka \textit{closed-loop calibration} approaches) or by observing the robot pose using visual sensors (\textit{open-loop calibration} approaches) (see \cite{Hollerbach2016}). Fig.~\ref{fig:schema_all_contacts} provides an overview. Next to traditional methods exploiting contact with the environment (e.g. robot touching a planar surface -- Fig.~\ref{fig:schema_all_contacts}B) or external metrology systems (e.g. laser trackers -- Fig.~\ref{fig:schema_all_contacts}D), we studied self-contact (Fig.~\ref{fig:schema_all_contacts}A) and self-observation (Fig.~\ref{fig:schema_all_contacts}C) as methods that are suited for automatic self-contained calibration.

An overview of the state of the art and future research directions regarding biologically inspired robot body models and self-calibration is provided in \cite{Hoffmann_EncRobotics_2021}, included in Appendix~\ref{appendixC:self-calibration}. In addition, four case studies are also included in this thesis. 

In \cite{Stepanova2019}, Appendix~\ref{appendixH:iCub_selfcalib}, we systematically studied on the simulated iCub humanoid robot how self-observation, self-contact, and their combination can be used for self-calibration. We found that employing multiple kinematic chains (self-observation and self-touch) is superior in terms of optimization results as well as observability. 

In \cite{Rozlivek_MultiRobotToolbox_2021}, Appendix~\ref{appendixI:multirobot}, we provided a unified formulation that makes it possible to combine traditional approaches with self-contained calibration available to humanoid robots in a single framework and single cost function. Second, we presented an open source toolbox for Matlab (\url{https://github.com/ctu-vras/multirobot-calibration}) that provides this functionality, along with additional tools for preprocessing (e.g., dataset visualization) and evaluation (e.g., observability/identifiability). 

In \cite{Rustler_NaoSkin_2021}, Appendix~\ref{appendixJ:nao_skin_calib}, we used self-contact as one of the methods to calibrate the positions of 970 pressure sensors on the body of a humanoid robot (the opposite approach to \cite{Roncone_ICRA_2014} where the skin was used to improve kinematic calibration). We experimentally compared the accuracy and effort associated with the following skin spatial calibration approaches and their combinations: (i) combining CAD models and skin layout in 2D, (ii) 3D reconstruction from images, (iii) using robot kinematics to calibrate skin by self-contact. 

Finally, in \cite{Stepanova2022}, Appendix~\ref{appendixK:motoman}, all four calibration methods schematically illustrated in Fig.~\ref{fig:schema_all_contacts}---self-contact, contact with a plane, self-observation, and external device---were experimentally compared on a dual-arm industrial manipulator.  The main findings were: (1) when applying the complementary calibration approaches in isolation, the self-contact approach yields the best and most stable results; (2) all combinations of more than one approach were always superior to using any single approach in terms of calibration errors and the observability of the estimated parameters. Combining more approaches delivers robot parameters that better generalize to the workspace parts not used for the calibration.

\chapter{Safe physical human-robot interaction and collaborative robots}
\label{chapter:pHRI}

Robots are leaving safety fences and start to share workspaces or even living spaces with humans. As they leave controlled environments and enter domains that are far less structured, they need to dynamically adapt to unpredictable interactions and guarantee safety at every moment. There has been rapid development in this regard in the last decade, with revisions of existing and introduction of new safety standards (\cite{ISO10218,ISO13855,ISO/TS15066}; see e.g. \cite{haddadin2016physical} for a survey)) and a rapidly growing market of collaborative robots. According to \cite{ISO/TS15066}, there are two ways of satisfying the safety requirements when a human physically collaborates with a robot: (i) \textit{Power and Force Limiting (PFL)} and (ii) \textit{Speed and Separation Monitoring (SSM)}.

For PFL, physical contacts with a moving robot are allowed but the forces / pressures / energy absorbed during a collision need to be within human body part specific limits. This translates onto lightweight structure, soft padding and no pinch points on the robot side, in combination with collision detection and response relying on motor load measurements, force/torque or joint torque sensing. This is addressed by interaction control methods for this post-impact phase (e.g., \cite{DeLuca2006,haddadin2008collision}; \cite{Haddadin2017} for a survey). The performance of robots complying with this safety requirement in terms of payload, speed, and repeatability is limited.

Safe collaborative operation according to speed and separation monitoring prohibits contacts with a moving robot and thus focuses on the pre-impact phase: a protective separation distance, $S_p$, between the operator and robot needs to be maintained at all times. When the distance decreases below $S_p$, the robot is commanded to halt. In industry, $S_p$ is typically safeguarded using light curtains (essentially electronic versions of physical fences) or safety-rated scanners that monitor 2D or 3D zones. However, the flexibility of such setups is limited---the information is reduced to detecting whether an object of a certain minimum volume has entered a predefined zone. The higher the robot kinetic energy, the bigger is its footprint on the factory floor. 

This topic ties with the rest of the thesis as follows. First, not only robot performance but also safety is dependent on its accuracy and calibration. Thus, machines learning and adapting their body models (Chapter~\ref{chapter:robots_learning_body_models} or self-calibrating robots in Chapter~\ref{chapter:self-calibration}) are also likely to be safer for their environments. Second, the use of artificial sensitive skins researched in my group can contribute an additional protective layer for collision detection and isolation during physical HRI (the PFL regime in particular). Third, adaptive multimodal representations of the robot peripersonal space (Chapter~\ref{chapter:robots_learning_pps}) could give rise to robots with a human-like margin of safety around their bodies and to new solutions to robot safety in the SSM regime.

In \cite{Svarny_IROS_2019}, Appendix~\ref{appendixL:SSM_PFL}, we deployed the two collaborative regimes (PFL and SSM) in a single application and studied the performance in a mock collaborative task under the individual regimes, including transitions between them. Additionally, we compared the performance under ``safety zone monitoring'' with keypoint pair-wise separation distance assessment relying on an RGB-D sensor and skeleton extraction algorithm to track human body parts in the workspace. Best performance has been achieved in the following setting: robot operates at full speed until a distance threshold between any robot and human body part is crossed; then, reduced robot speed per power and force limiting is triggered. Robot is halted only when the operator's head crosses a predefined distance from selected robot parts.

In \cite{svarny_3Dcollision_2021}, Appendix~\ref{appendixM:3dCFM}, we measured the forces exerted by two collaborative manipulators moving downward against an impact measuring device. First, we empirically showed that the impact forces can vary by more than 100 percent within the robot workspace. The forces are negatively correlated with the distance from the robot base and the height in the workspace. Second, we presented a data-driven model, 3D Collision-Force-Map, predicting impact forces from distance, height, and velocity and demonstrate that it can be trained on a limited number of data points. Third, we analyzed the force evolution upon impact and found that clamping never occurs for one of the robots (UR10e). We showed that formulas relating robot mass, velocity, and impact forces from ISO/TS 15066 \cite{ISO/TS15066} are insufficient---leading both to significant underestimation and overestimation and thus to unnecessarily long cycle times or even dangerous applications. We proposed an empirical method that can be deployed to quickly determine the optimal speed and position where a task can be safely performed with maximum efficiency.

Finally, in collaboration with the group of Sami Haddadin at Technical University of Munich, we investigated how the velocity of a collaborative manipulator can be modulated by monitoring the interaction with a human. The details are in \cite{Zardykhan_Humanoids_2019} (not included in Appendix).

\bibliographystyle{apalike}
\bibliography{Hoffmann_habilitation}



\backmatter

\end{document}

%% file: ifi_pages.tex
\begin{titlepage}
 
\begin{centering}
\begin{large} Czech Technical University in Prague \end{large} \\
\begin{large} Faculty of Electrical Engineering \end{large} \\
\begin{large} Department of Cybernetics \end{large}\\ [0.5cm]


\includegraphics[width=0.3\textwidth]{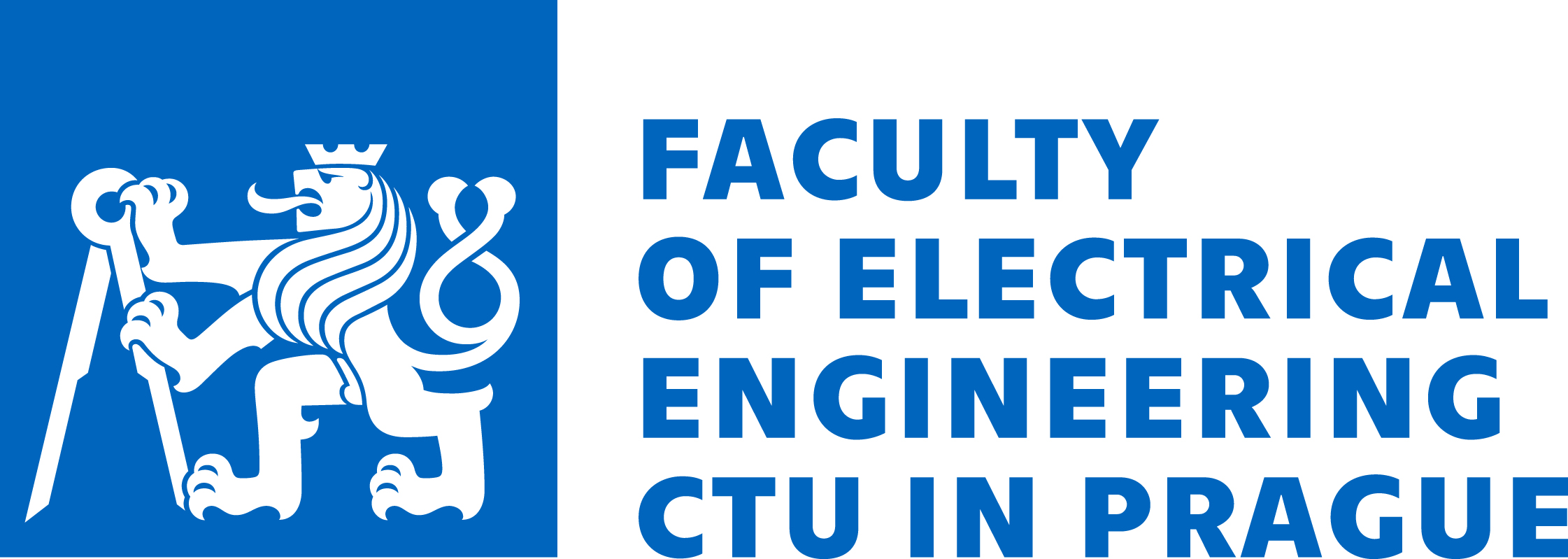} \\ [3cm]

{\huge \bfseries Learning Body Models:}\\
{\huge \bfseries From Humans to Humanoids} \\[3cm]


{\large Habilitation Thesis} \\[5cm]



{\Large \bfseries Mat\v{e}j Hoffmann} \\[0.5cm] 

\begin{large}Prague, November 2021\end{large} \\[1.5cm]



\end{centering}







\end{titlepage}